\documentclass[10pt,twocolumn,letterpaper]{article}

\usepackage[pagenumbers]{iccv} % To force page numbers, e.g. for an arXiv version

\usepackage[accsupp]{axessibility}  % Improves PDF readability for those with disabilities.
\usepackage[pagebackref=true,breaklinks=true,letterpaper=true,colorlinks,bookmarks=false]{hyperref}

\usepackage{algorithm}      % http://ctan.org/pkg/algorithms
\usepackage{algpseudocode}  % http://ctan.org/pkg/algorithmicx
\usepackage{multirow} % Required for multiple rows
\usepackage{multicol} % Required for multiple columns
\usepackage{xcolor}
\usepackage{colortbl}
\definecolor{orange}{rgb}{1.0, 0.6, 0.2}
\usepackage{verbatim}
\usepackage{graphicx}
\usepackage{caption}
\usepackage{wrapfig}
\definecolor{darkyellow}{rgb}{0.85, 0.65, 0.0}

\def\paperID{1326} % *** Enter the Paper ID here
\def\confName{ICCV}
\def\confYear{2025}

\begin{document}
\title{PerLDiff: Controllable Street View Synthesis Using Perspective-Layout Diffusion Model}

%%%%%%%%% AUTHORS - PLEASE UPDATE
\author{
\hspace{-0.5cm}
Jinhua Zhang\textsuperscript{1}\footnotemark[1]\quad Hualian Sheng\textsuperscript{2}\footnotemark[1]\quad
Sijia Cai\textsuperscript{2}\quad Bing Deng\textsuperscript{2}\quad Qiao Liang\textsuperscript{2}\quad \\ Wen Li \textsuperscript{1}\quad Ying Fu\textsuperscript{3}\quad Jieping Ye\textsuperscript{2}\quad Shuhang Gu\textsuperscript{1}\footnotemark[2]\\%\Letter
\hspace{-0.5cm}
\small \textsuperscript{1}{University of Electronic Science and Technology of China} \hspace{0pt}\quad
\small \textsuperscript{2}{Independent Researcher}\quad \small \textsuperscript{3}{Beijing Institute of Technology} \hspace{0pt}\\
\hspace{-0.5cm}
{\tt \small \{jinhua.zjh, shenghualian.shl, shuhanggu\}@gmail.com}\\
\small Codes and models: \url{https://github.com/LabShuHangGU/PerLDiff}
}

\maketitle

\renewcommand{\thefootnote}{\fnsymbol{footnote}} 
\footnotetext[1]{Equal contribution.}
\footnotetext[2]{Corresponding author.} 

\begin{abstract}
Controllable generation is considered a potentially vital approach to address the challenge of annotating 3D data, and the precision of such controllable generation becomes particularly imperative in the context of data production for autonomous driving. 
Existing methods focus on the integration of diverse generative information into controlling inputs, utilizing frameworks such as GLIGEN or ControlNet, to produce commendable outcomes in controllable generation. 
However, such approaches intrinsically restrict generation performance to the learning capacities of predefined network architectures.  
In this paper, we explore the innovative integration of controlling information and introduce PerLDiff (\textbf{Per}spective-\textbf{L}ayout \textbf{Diff}usion Models), a novel method for effective street view image generation that fully leverages perspective 3D geometric information.
Our PerLDiff employs 3D geometric priors to guide the generation of street view images with precise object-level control within the network learning process, resulting in a more robust and controllable output. 
Moreover, it demonstrates superior controllability compared to alternative layout control methods. Empirical results justify that our PerLDiff markedly enhances the precision of controllable generation on the NuScenes and KITTI datasets.
% 
% Code is available at \small \url{https://github.com/LabShuHangGU/PerLDiff}.
\end{abstract}

\section{Introduction}
\label{sec:intro}
The advancement of secure autonomous driving systems is fundamentally dependent on the accurate perception of the vehicle's environment.  
Recently, perception utilizing Bird's Eye View (BEV) has seen rapid progress, markedly pushing forward areas such as 3D object detection~\cite{li2022bevformer,liu2023bevfusion} and BEV segmentation~\cite{zhou2022cross}.
Nonetheless, these systems necessitate extensive datasets with high-quality 3D annotations, the acquisition of which typically involves two consecutive steps: data scene collection and subsequent labeling. Each of these steps incurs significant expenses and presents considerable challenges in terms of data acquisition.

\begin{figure}[t]
    \centering
    % \hfill
    \includegraphics[width=0.95\linewidth]{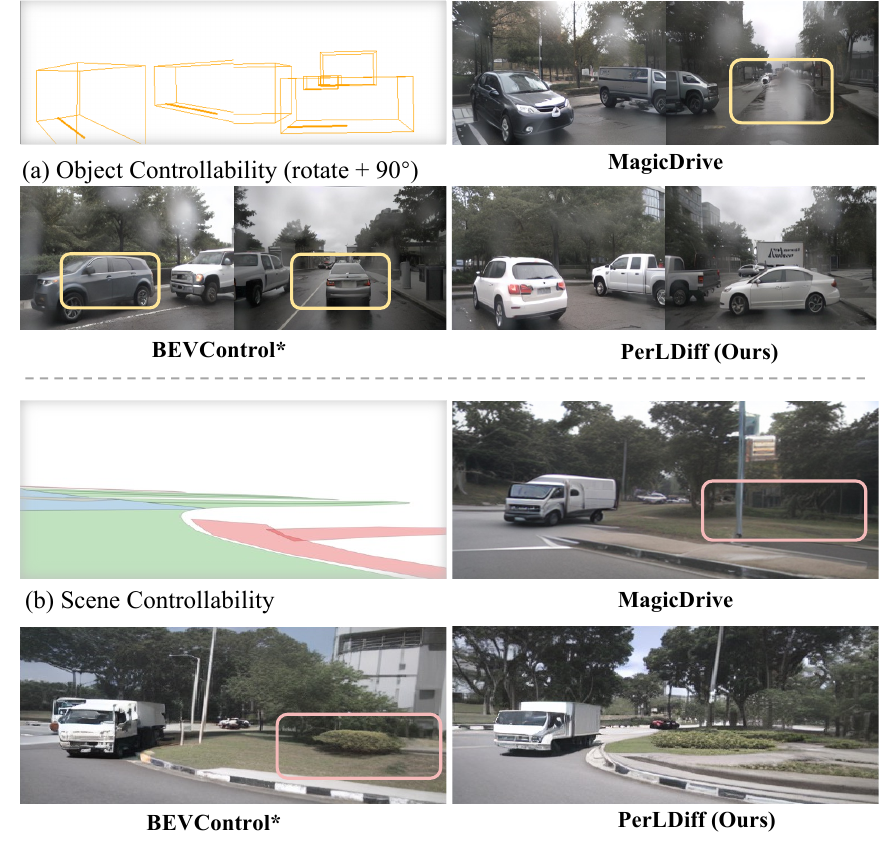} 
   \caption{PerLDiff enhances controllability over BEVControl* and MagicDrive using geometric priors. \textit{\textcolor{darkyellow}{Top}}: Demonstrates object controllability by adjusting the 3D annotation yaw by 90 degrees. \textit{\textcolor{pink}{Bottom}}: Shows scene controllability through the alignment of the street map with the generated image. Regions highlighted by rectangles indicate areas where the generated images fail to achieve control and alignment with ground truth conditions.}
   \label{fig:intro}
   \vspace{-1em}
\end{figure}

To mitigate issue of data scarcity, the adoption of generative technologies~\cite{van2017neural, esser2021taming, ho2020denoising} has proven practical for reversing the order of data annotation.
The paradigm of this approach is to use the collected annotation as controlling information to generate the corresponding lifelike images depicting urban street scenes.
By implementing this strategy, it is possible to dramatically lower the costs associated with data annotation while also facilitating the generation of extensive long-tail datasets, subsequently leading to improvements in the perception model's performance.
Pioneering research, exemplified by BEVGen~\cite{swerdlow2024street}, harnesses the capabilities of autoregressive transformers~\cite{van2017neural, esser2021taming} to render detailed visualizations of street scenes. 
Additionally, subsequent studies such as BEVControl~\cite{yang2023bevcontrol} and MagicDrive~\cite{gao2023magicdrive} employ diffusion-based techniques, including GLIGEN~\cite{li2023gligen} and ControlNet~\cite{zhang2023adding}, to integrate controlling information through a basic cross-attention mechanism.
However, these methods simply extract integrated conditional features from controlling information to adjust the generation process and are limited in making full use of detailed geometric layout information for accurate attention map manipulation.
While these techniques make strides towards meeting the requirements toward generating data from 3D annotations, Fig.~\ref{fig:intro} demonstrates that there remains significant potential for improvement in scene and object controllability.

To this end, in this paper, we introduce the perspective-layout diffusion models (PerLDiff), a novel method specifically designed to enable precise control over street view image generation at the object level.
In addition to extracting integrated conditional features from controlling condition information, \textit{i.e.} 3D annotations, our PerLDiff explicitly renders perspective layout masking maps as geometric priors.
Subsequently, a PerL-based controlling module (PerL-CM) is proposed to leverage the geometric priors, \textit{i.e.} perspective layout masking maps. Within PerL-CM, an innovative PerL-based cross-attention mechanism is utilized to accurately guide the generation of each object with their corresponding condition information.
We integrate PerL-CM into the pre-trained Stable Diffusion model~\cite{rombach2022high} and fine-tune it on our training dataset.
Consequently, our PerLDiff incorporates the formidable generative capabilities of Stable Diffusion with the finely detailed geometric priors of perspective layouts, effectively harnessing their respective strengths for precise object-level image synthesis.
Overall, our PerLDiff is capable of generating precise, controllable street view images while also maintaining high fidelity (see Section~\ref{sec:controlled_generation_based_on_3d_geometric_priors} for details).

The main contributions of this paper are summarized as following three-folds:
\begin{itemize}
\item We present PerLDiff, a newly developed framework crafted to synthesize street view images from user-defined 3D annotations. Our PerLDiff carefully orchestrates the image generation process at the object level by leveraging perspective layout masks as geometric priors.

\item We propose a PerL-based cross-attention mechanism that utilizes perspective layout masking maps from 3D annotations to enhance the underlying cross-attention mechanism within PerL-CM. This method enables precise control over the street view image generation process by integrating road geometry and object-specific information derived from 3D annotations.

\item Our PerLDiff method attains state-of-the-art performance on the NuScenes~\cite{caesar2020nuscenes} and KITTI~\cite{Geiger2012CVPR} dataset compared to existing methods, markedly enhancing detection and segmentation outcomes for synthetic street view images. Furthermore, it holds the potential to function as a robust traffic simulator in the future.
\end{itemize}

\section{Related Work}
\label{relate_work}
% \subsection{Diffusion-based Generative Models in Image Synthesis}
\textbf{Diffusion-based Generative Models in Image Synthesis.}
Initially developed as a method for modeling data distributions through a sequence of Markov chain diffusion steps~\cite{sohl2015deep, song2020score, song2020sliced}, diffusion models have undergone rapid advancement.
Ho et al.~\cite{ho2020denoising} introduced denoising diffusion probabilistic models (DDPMs), which have established new benchmarks in the quality of image synthesis. Following efforts have aimed to enhance the efficiency and output diversity of these models by investigating various conditioning strategies~\cite{dhariwal2021diffusion, choi2021ilvr}, architectural adjustments, and training methodologies to refine the image synthesis process~\cite{hong2023improving}.
Nichol and Dhariwal~\cite{dhariwal2021diffusion} proved that diffusion models can be text-conditioned to produce coherent images that are contextually appropriate.
Furthermore, advances such as multimodal-conditioned diffusion models have effectively utilized layout images ~\cite{li2023gligen,zhang2023adding,rombach2022high,qu2023layoutllm}, semantic segmentation maps ~\cite{li2023gligen,zhang2023adding}, object sketches ~\cite{voynov2023sketch,mou2023t2i,li2023gligen,zhang2023adding}, and depth maps ~\cite{mou2023t2i,li2023gligen,zhang2023adding} to inform the generative process. These methods enable more targeted manipulation of the imagery, thus yielding complex scenes characterized by enhanced structural integrity and contextual pertinence.
\begin{figure*}[htbp]
    \centering
    \includegraphics[width=0.95\textwidth]{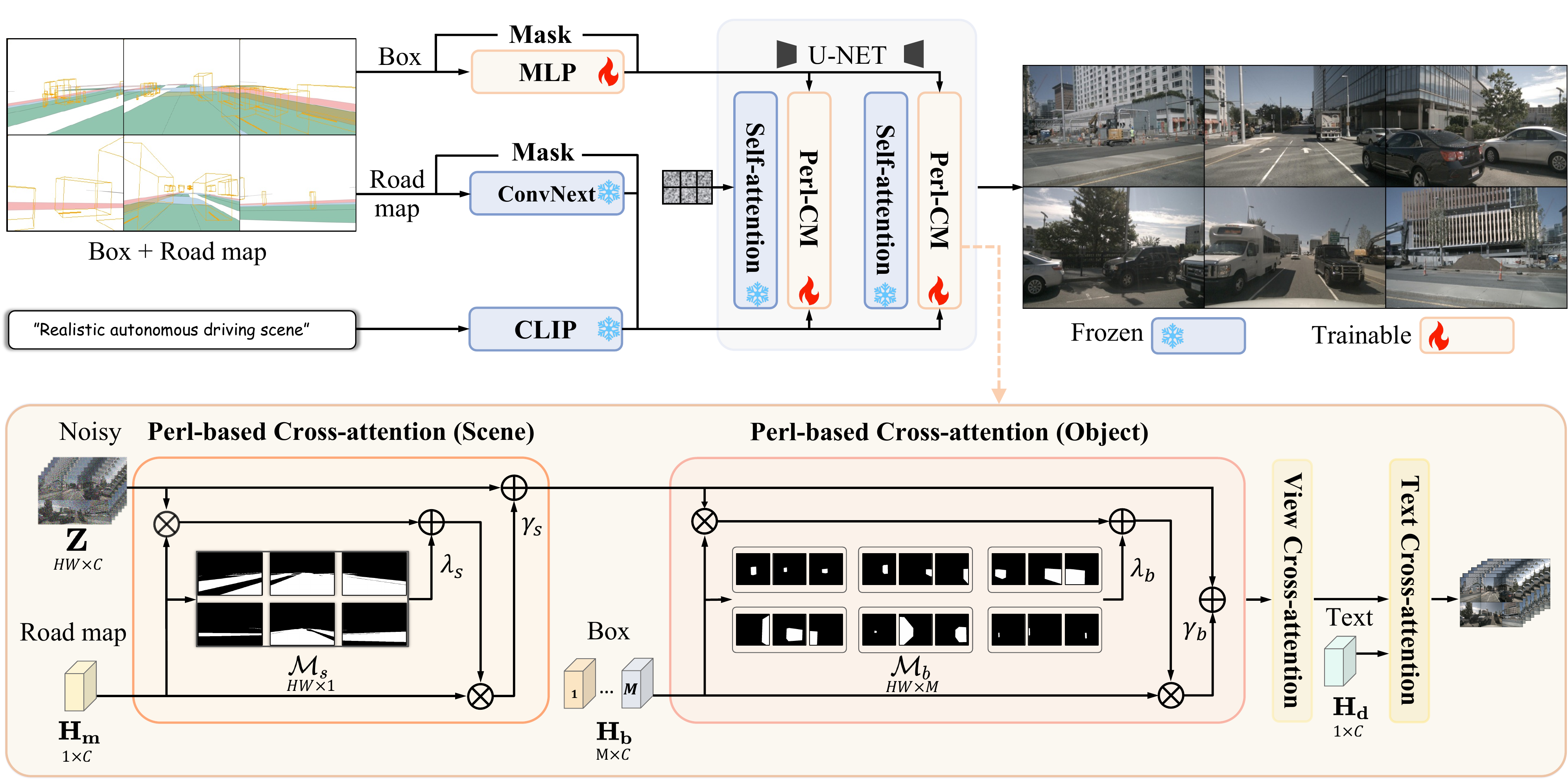} 
    \caption{Overview of \textbf{PerLDiff} framework for multi-view street image generation (multi-view dimension omitted for simplicity). PerLDiff utilizes perspective layout masking maps derived from 3D annotations to integrate scene information and object bounding boxes. \textbf{PerL-CM} is responsible for integrating control information through employing \textbf{PerL-based cross-attention (Scene \& Object)} mechanism, using PerL masking map (road \& box) as geometric priors to guide object-level image generation with high precision. View cross-attention ensures consistency across multiple views, while Text cross-attention integrates textual scene descriptions to facilitate further adjustments.}
    \label{fig:PerLDiff_framwork}
    \vspace{-1em}
\end{figure*}
%

% \subsection{Data Generation for Autonomous Driving}
\noindent\textbf{Data Generation for Autonomous Driving.} 
BEVGen~\cite{swerdlow2024street} represents the pioneering endeavor to harness an autoregressive Transformer ~\cite{van2017neural, esser2021taming} for synthesizing multi-view images pertinent to autonomous driving.
Building upon this, BEVControl~\cite{yang2023bevcontrol} introduces a novel method that incorporates a diffusion model~\cite{ho2020denoising} for street view image generation, and integrates cross-view attention mechanisms to maintain spatial coherence across neighboring camera views.
Subsequently, MagicDrive~\cite{gao2023magicdrive} propels the field forward by refining the method for controlling input conditions, drawing insights from ControlNet~\cite{zhang2023adding}.
DrivingDiffusion~\cite{li2023drivingdiffusion} further augments the framework by introducing a consistency loss designed to achieve the perceptual uniformity requisite for high precision in the generation of video from autonomous driving.
Panacea~\cite{wen2023panacea} broadens the capabilities of the model by tackling the challenge of ensuring temporal consistency in video.
In contrast to the above approaches, which primarily utilize controlling input conditions to steer the image generation process, our PerLDiff exploits detailed geometric layout information from the input to directly guide object generation with higher precision.

% \subsection{Geometric Constraints in Image Generation}
\noindent\textbf{Geometric Constraints in Image Generation.}
Incorporating geometric priors into image synthesis has been explored to a lesser extent. Work on 3D-aware image generation~\cite{nguyen2019hologan, niemeyer2021giraffe} suggests the feasibility of integrating geometric information into generative processes to improve spatial coherence. Nevertheless, these methods often rely on complex 3D representations and may not be directly applied to diffusion model frameworks. 
Recently, BoxDiff~\cite{xie2023boxdiff} reveals a spatial correspondence between the attention map produced by the diffusion model and the corresponding generated image. During the testing phase, the geometric configuration of the attention map's response values is adjusted to yield a more precise image generation.
ZestGuide~\cite{couairon2023zero} introduces a loss function that enforces a geometric projection onto the attention map, further refining the shape of the attention map's response values to closely approximate the geometric projection of the control information during the inference stage.
However, utilizing text prompts to facilitate the generation of complex urban environment layouts poses inherent challenges, owing to the need for crafting intricate prompts to accurately depict urban environments.
Furthermore, modifying the cross-attention map to impose strict constraints during the denoising phase of inference can disrupt the intrinsic relationships, leading to a suboptimal approach to synthesizing controllable images.
In contrast, PerLDiff incorporates geometric prerequisites as training priors to guide the generation of street view images, offering a more effective solution.
%
%
% \vspace{-2.1em}
\section{Controllable Street View Generation Based on Perspective Layout}
\label{sec:controlled_generation_based_on_3d_geometric_priors}
In this paper, we introduce PerLDiff, depicted in the Fig.~\ref{fig:PerLDiff_framwork}, which is designed to enable controllable multi-view street scene image generation using 3D annotations.
Specifically, our PerLDiff leverages perspective projection information from 3D annotations as controlling condition within the training regimen and utilizes perspective layout masks as geometric priors, enabling accurate guidance in object generation.
In the following sections, we delineate the process of encoding the controlling condition information from the 3D annotations in the Section~\ref{subsec:CIE}. Additionally, we explain how incorporating the perspective layout knowledge to ensure scene and object controllability in street view image generation in the Section~\ref{subsec:enhancing-control}.

\subsection{Controlling Conditions Encoding}
\label{subsec:CIE}
Given 3D annotations of a street scene, our goal is to generate multi-view street images. 
To be more specific, for controllable street view image generation, we extract not only scene information (\textit{i.e.}, textual scene descriptions $\mathbf{S}_d$ and street maps $\mathbf{M}$ revealing features such as road markings and obstacles) but also object information (\textit{i.e.}, bounding box parameters $\mathbf{P}$ and the associated object category $\mathbf{Y}$) from 3D annotations as controlling conditions. 
These controlling conditions encompass rich semantic and geometric information, so establishing a robust encoding method to utilize this information is essential for generating street view images. Hereafter, we present our controlling condition encoding approach. 
For simplicity, we omit the details of multi-view perspectives.

% \subsubsection{PerL Scene Information}
\noindent\textbf{PerL Scene Information}
encompasses perspective scene images and supplemental data specific for the whole scene. 
Typically, the selected scene for annotation is coupled with a street map of the driving environment, which visually differentiates between the road and other background elements using distinct colors.
In addition, a generic textual description of the scene is customizable to align with particular scenarios. 
We employ ConvNext~\cite{liu2022convnet} and the CLIP text encoder~\cite{radford2021learning} to encode the perspective road map, denoted as $\mathbf{S}_m$, derived from the projection of the street map and the textual scene description $\mathbf{S}_d$, respectively.
This approach results in the generation of encoded scene features $\mathbf{H}_m \in \mathbb{R}^{1 \times C}$ for the road map and $\mathbf{H}_d \in \mathbb{R}^{1 \times C}$ for the textual scene description:
\begin{align}
    \mathbf{H}_m = \text{ConvNext}(\mathbf{S}_m),\qquad \qquad
    \mathbf{H}_d = \varphi(\mathbf{S}_d).
\end{align}
%
% \subsubsection{PerL Object Information}
\noindent\textbf{PerL Object Information} 
encapsulates perspective geometric data alongside object category information, which stems from the projections of annotated 3D boxes. 
Through projecting 3D annotations onto their corresponding perspective images, we ascertain eight 2D corner points for each bounding box within a single image, denoted as $\mathbf{P}_{g} \in \mathbb{R}^{M \times 2 \times 8}$, where $M$ represents the maximum number of bounding boxes and eight corresponds to the number of corners per bounding box.
In conjunction with the object's categorical text $\mathbf{P}_{c} = \{\mathbf{P}_{c_{i}}\}_{i=1}^{M} $, we derive the encoded box geometric features $\mathbf{H}_g \in \mathbb{R}^{M \times C}$ and the box categorical features $\mathbf{H}_c \in \mathbb{R}^{M \times C}$, which are illustrated as follows:
\begin{align}\label{eq:fourier_encoder}
\mathbf{H}_g = \mathcal{F}(\mathbf{P}_g),\qquad \qquad 
\mathbf{H}_c = \varphi(\mathbf{P}_c),
\end{align}
where $\mathcal{F}(\cdot)$ is the Fourier embedding function~\cite{mildenhall2021nerf}, $\varphi(\cdot)$ represents the pre-trained text embedding encoder of CLIP~\cite{radford2021learning} and $C$ representing the dimension of features. 
Furthermore, we concatenate the encoded geometric features $\mathbf{H}_g$ and categorical features $\mathbf{H}_c$ and subsequently pass the concatenated vector through a Multilayer Perceptron (MLP)~\cite{taud2018multilayer} $\mathbf{F}$, to achieve feature fusion $\mathbf{H}_b \in \mathbb{R}^{M \times C}$. The resulting fused box feature representation is given by:
\begin{align}
\mathbf{H}_b = \mathbf{F}([\mathbf{H}_g, \mathbf{H}_c]).
\end{align}
We subsequently input the encoded conditions into the denoising diffusion model to guide the generation process. This is achieved utilizing PerL-based cross-attention that incorporates PerL masking maps, as detailed below.
\begin{figure*}[t]
    \centering
    \includegraphics[width=\textwidth]{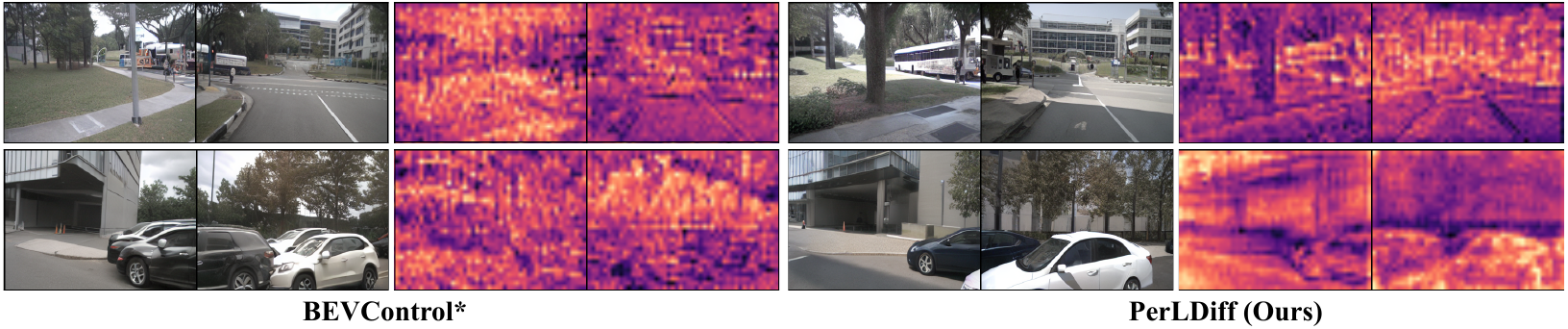} 
    \caption{Visualization of cross-attention maps reveals perceptual congruency with the generated image. BEVControl* produces disorganized and vague attention maps, which result in inferior image quality. Conversely, our PerLDiff method fine-tunes the response within the attention maps, resulting in more accurate control information at the object level and improved image quality. Please see more qualitative examples in the Appendix.}
    \label{fig:pbca_vs_base_atten_map_1}
    \vspace{-1em}
\end{figure*}

\subsection{Object Controllability via PerL-based Controlling Module}
\label{subsec:enhancing-control}
PerL-based Controlling Module (PerL-CM) is responsible for integrating controlling condition information, which encompasses both the PerL scene and object information, into the latent feature maps of noisy street view images. 
This integration is primarily achieved via the scene and object \textbf{PerL-based cross-attention} mechanism. 
Initially, this mechanism assigns initial values to the attention maps, under the guidance of road and box PerL masking maps. 
Throughout the training of the network, these values are optimized to ensure that the response of the attention map accurately corresponds to the regions where the objects are located.
Subsequently, the information from both the road map and the bounding box are sequentially integrated into the noise street view image.
To more effectively integrate PerL scene and object information, a gating operation  similar to GLIGEN~\cite{li2023gligen} is used, 
which dynamically adjusts the contribution of condition information according to the adaptive process. 
To ensure multi-view consistency, \textbf{View Cross-attention} leverages information from the immediate left and right views for uniformity across various perspectives. 
Additionally, \textbf{Text Cross-attention} manipulates the weather and lighting conditions of street scenes using textual scene description. 
The details of the PerL-CM process are delineated in the Appendix.
%

% \subsubsection{PerL Masking Map (road \& box)}
\noindent\textbf{PerL Masking Map (road \& box)} 
is comprised of PerL road masking map $\mathcal{M}_s \in \mathbb{R}^{HW \times 1}$ and PerL box masking map $\mathcal{M}_b \in \mathbb{R}^{HW \times M}$ , where $H$ and $W$ represent the height and width dimensions of the image, respectively. These masking maps are articulated as follows:
\begin{align}
\mathcal{M}_{s} = \Upsilon(\mathbf{S}_m),\quad \mathcal{M}_{b} = \Phi(\mathbf{P}_g),
\end{align}
where $\Upsilon(\cdot)$ generates the masking map for the non-empty regions of the projected road maps. Meanwhile, $\Phi(\cdot)$ produces the masking map corresponding to the inner region of each projected 3D bounding box for every perspective image, enabling precise control at the object level.

\noindent\textbf{PerL-based Cross-attention (Scene \& Object)} 
% \subsubsection{PerL-based Cross-attention (Scene \& Object)}
leverages the prior masking maps to enhance the learning of cross-attention between the input controlling conditions and the noisy street view images. 
As depicted in Fig.~\ref{fig:pbca_vs_base_atten_map_1}, the cross-attention map exhibits perceptual equivalence with the generated street view image. However, this correspondence is imprecise and lacks the necessary alignment during the training stage.
To this end, our approach utilizes a PerL-based cross-attention mechanism that incorporates geometric knowledge derived from both the scene context and the bounding box into the computation of the cross-attention map. 
In PerLDiff, the road map and object bounding box data are seamlessly merged with the noisy street view images throughout each stage of the denoising process.
For the sake of notational simplicity, the linear embeddings and normalization steps typically involved in the attention mechanism have been omitted.
\begin{align}
    \mathcal{A}_s = \textit{softmax}(\lambda_s \cdot \mathcal{M}_s + \mathbf{Z}\mathbf{H}_m^{T}/\sqrt{d}),\\ \mathcal{A}_b = \textit{softmax}(\lambda_b \cdot \mathcal{M}_b + \mathbf{Z_s}\mathbf{H}_b^{T}/\sqrt{d}),
\end{align}
where $\lambda_s$ and $\lambda_b$ are weight parameters that control the influence of the masking map, $d$ denotes the dimensionality, and $\mathbf{Z}$, $\mathbf{Z}_b$, $\mathbf{Z}_s \in \mathbb{R}^{HW \times C}$ represent different noisy street view images. 
Here, $\mathcal{A}_s \in \mathbb{R}^{HW \times 1} $ characterizes the association between the road map and the noisy image, while $\mathcal{A}_b \in \mathbb{R}^{HW \times M}$ clarifies the relationship between the conditions of the object bounding box and the noisy image.
For visualization purposes, as shown in Fig.~\ref{fig:pbca_vs_base_atten_map_1}, we average $\mathcal{A}_b$ along the second dimension and merge all object attention maps into one representation at step 50 of the DDIM process in the last block of U-Net.
The final noisy street view image is synthesized through an attention-based aggregation mechanism enhanced by a residual connection, which can be expressed as:
\begin{align}
    \mathbf{Z_s} = \gamma_s \cdot \mathcal{A}_s\mathbf{H}_m + \mathbf{Z}, \qquad 
    \mathbf{Z_b} = \gamma_b \cdot \mathcal{A}_b\mathbf{H}_b + \mathbf{Z_s},
\end{align}
where $\gamma_s$ and $\gamma_b$ represent learnable parameters modulating the influence of respective conditions.
%

% \subsubsection{View Cross-attention}
\noindent\textbf{View Cross-attention} is corroborated by preliminary works such as BEVControl~\cite{yang2023bevcontrol}, MagicDrive~\cite{gao2023magicdrive}, DrivingDiffusion~\cite{li2023drivingdiffusion}, and Panacea~\cite{wen2023panacea}, plays an instrumental role in facilitating the synthesis of images that maintain visual consistency across varying camera perspectives. For additional information, please see the Appendix.

\noindent\textbf{Text Cross-attention} enhances Stable Diffusion~\cite{rombach2022high} model's ability to modulate street scenes through textual scene description. This capability is crucial for dynamically adapting the rendering of street scenes to accommodate various lighting and weather conditions. By integrating detailed textual scene description, PerLDiff can alter visual elements such as illumination and atmospheric effects, ensuring that the generated images reflect the specified conditions. For qualitative examples, see the the Appendix.

% \vspace{-0.6em}
\subsection{Discussion}
\label{subsec:discussion}
In contrast to previous approaches for autonomous driving such as BEVControl~\cite{yang2023bevcontrol}, MagicDrive~\cite{gao2023magicdrive}, DrivingDiffusion~\cite{li2023drivingdiffusion}, and Panacea~\cite{wen2023panacea}, which employ a basic cross-attention mechanism to integrate controlling condition information, our PerLDiff leverages geometric priors via PerL masking map. This approach directs the generation of each object with its respective control information during the training phase, effectively countering the common misalignment between the attention map and condition information that often results in compromised image controllability.
For instance, the attention map of BEVControl~\cite{yang2023bevcontrol}, illustrated in Fig.~\ref{fig:pbca_vs_base_atten_map_1}, demonstrates disorganized patterns and lacks precision.
Conversely, our PerLDiff markedly enhances the accuracy of generated images and the granularity of condition information at the object level by ensuring meticulous guidance within the attention map. For more qualitative results, please see the Appendix.

\section{Experiments}
\label{sec:experiments}
We assess PerLDiff's ability to control quality across multiple benchmarks, including multi-view 3D object detection~\cite{li2022bevformer,liu2023bevfusion,zhang2021objects,wang2023exploring}, BEV segmentation~\cite{zhou2022cross} and monocular 3D object detection~\cite{zhang2021objects}. Subsequently, we conduct ablation studies to ascertain the contribution of each component within our proposed methodology.

\subsection{Datasets}
\noindent\textbf{NuScenes} dataset comprises 1,000 urban street scenes, traditionally segmented into 700 for training, 150 for validation, and 150 for testing. Each scene features six high-resolution images (900$\times$1600), which together provide a complete 360-degree panoramic view of the surroundings. Additionally, NuScenes includes comprehensive road maps of the driving environment, featuring details such as lane markings and obstacles. 
We extend the class and road type annotations similar to MagicDrive~\cite{gao2023magicdrive} and NuScenes, incorporating ten object classes and eight road types for map rendering. To address the resolution limitations of the U-Net architecture in Stable Diffusion~\cite{ronneberger2015u}, we adopt image resolutions of 256$\times$384 as in BEVFormer~\cite{li2022bevformer}, and 256$\times$704 following BEVFusion (Camera-Only)~\cite{liu2023bevfusion}.

\noindent\textbf{KITTI} dataset contains 3,712 images for training and 3,769 images for validation. KITTI has only one perspective image and does not have road map information. Given the varied image resolutions in KITTI (approximately 375$\times$1242), we pad them to 384$\times$1280 for generative learning.
\subsection{Main Results}
In this subsection, we assess PerLDiff's generative quality through the perception results of several pre-trained methods: BEVFormer~\cite{li2022bevformer}, BEVFusion (Camera-Only)~\cite{liu2023bevfusion}, and StreamPETR~\cite{wang2023exploring} for multi-view 3D detection; CVT~\cite{zhou2022cross} for BEV segmentation; and MonoFlex~\cite{zhang2021objects} for monocular 3D detection trained on the KITTI.
Additionally, we leverage synthesized dataset to enhance the performance of various 3D detection models (\textit{i.e.}, BEVFormer and StreamPETR) on the NuScenes \textit{test}, validating the effectiveness of our PerLDiff.

\begin{table}[t]
    \centering
    \resizebox{\linewidth}{!}{
    \begin{tabular}{@{}lcc|ccc@{}}
    \toprule
    Method & Detector & FID$\downarrow$ & mAP$\uparrow$ & NDS$\uparrow$ & mAOE$\downarrow$ \\
    \midrule
    Oracle & BEVFormer & -- & 27.06 & 41.89 & 0.54   \\
    Oracle & BEVFusion & -- & 35.54 & 41.20 & 0.56   \\
    \midrule
    MagicDrive~\cite{gao2023magicdrive} & BEVFusion & 16.20 & 12.30 & 23.32 & -- \\
    MagicDrive* & BEVFusion & 15.92 & 10.27 & 20.42 & 0.78  \\
    BEVControl* & BEVFusion & \textbf{13.05}  & 9.98 & 19.61 & 0.94  \\
    \rowcolor{gray!25}PerLDiff (Ours) & BEVFusion & 13.36  & \textbf{15.24} & \textbf{24.05} & \textbf{0.78} \\
    \midrule
    BEVGen \citep{swerdlow2024street} & --  & 25.54 & -- & -- & --  \\
    BEVControl~\cite{yang2023bevcontrol} & BEVFormer & 24.85 & 19.64 & 28.68 & 0.78  \\
    BEVControl* & BEVFormer & 13.05  & 16.48 & 28.08 & 0.88  \\
    MagicDrive* & BEVFormer & 15.92 & 15.21 & 28.79 & 0.81  \\
    \rowcolor{gray!25}PerLDiff (Ours) & BEVFormer & 13.36  & \textbf{25.10} & \textbf{36.24} & \textbf{0.72} \\
    \bottomrule
    \end{tabular}
    }
    \caption{Comparison of the controllability on NuScenes \textit{validation} set. 
    BEVControl*, serving as the baseline, employs standard cross-attention mechanisms contrary to PerL-based cross-attention utilized in PerLDiff. MagicDrive* represents our replication using the official configuration. \textbf{Bold} indicates the best result.
    }
    \label{table:bev_detection}
     \vspace{-1em}
\end{table}

% \noindent\textbf{Controllable Generation on NuScenes.} 
\subsubsection{Controllable Generation on NuScenes}
To evaluate the effectiveness of PerLDiff, we trained the model on the NuScenes \textit{train} set and subsequently generated a synthetic \textit{validation} set using the provided road maps and 3D annotations. The controllability of PerLDiff was examined by applying perception models, originally trained on the real \textit{train} set, to our synthetic \textit{validation} set. As summarized in Tab.~\ref{table:bev_detection}, PerLDiff outperforms competing methods across most metrics, as tested with BEVFormer~\cite{li2022bevformer} and BEVFusion~\cite{liu2023bevfusion}. In a rigorous comparison, we replicated BEVControl~\cite{yang2023bevcontrol} using identical settings, with the exception of our innovative element: the PerL-based cross-attention mechanism. PerLDiff demonstrates notable improvements, with increases of 8.62\%, 8.16\%, and 0.16\% in mean Average Precision (mAP), NuScenes Detection Score (NDS), and mean Average Orientation Error (mAOE), respectively, compared to BEVControl* when using BEVFormer. With BEVFusion, it achieves gains of 5.26\%, 4.44\%, and 0.16\% in these metrics against BEVControl*, confirming its effectiveness at a resolution of $256\times384$. 

\begin{table}[t]
    \centering
    \resizebox{\linewidth}{!}{
    \begin{tabular}{@{}lc|ccccc@{}}
    \toprule
    Method & Oracle & BEVGen & BEVControl &BEVControl* &MagicDrive*&PerLDiff (Ours) \\
    \midrule
    Road mIoU$\uparrow$ & 70.35 & 50.20 &  60.80 & 60.74 &  55.56 & \textbf{61.26}\\
    Vehicle mIoU$\uparrow$ & 33.36 & 5.89 & 26.86 & 22.47& 24.81& \textbf{27.13}\\
    \bottomrule
    \end{tabular}
    }
    \caption{Comparison of the controllability performance on the NuScenes \textit{validation} set using BEV segmentation metrics in CVT~\cite{zhou2022cross}. \textbf{Bold} indicates the best result.}
     \vspace{-1em}
    \label{table:cvt_seg}
\end{table}
\begin{table}[t]
    \centering
    % \vspace{-1em}
    \resizebox{\linewidth}{!}{
    \begin{tabular}{@{}lcccccccc@{}}
    \toprule
    \multirow{2}{*}{Method}    & \multicolumn{4}{c}{KITTI} & \multicolumn{4}{c}{NuScenes $\rightarrow$ \text{KITTI}} \\ 
    \cmidrule(lr){2-5}\cmidrule(lr){6-9}
    & Easy$\uparrow$ & Mod.$\uparrow$ & Hard$\uparrow$ & FID$\downarrow$ & Easy$\uparrow$ & Mod.$\uparrow$ & Hard$\uparrow$ & FID$\downarrow$ \\ \midrule
    Oracle & 22.29 & 15.54 & 13.38 & -- &  22.29 & 15.54 & 13.38 & -- \\
    BEVControl* & 0.33 & 0.29 & 0.39 & 39.47 & 1.32 & 1.51 & 1.64 & 32.96 \\
    \rowcolor{gray!25}PerLDiff (Ours) & \textbf{11.04} & \textbf{7.44} & \textbf{6.03} & \textbf{39.03} & \textbf{13.12} & \textbf{9.24} & \textbf{7.59} & \textbf{31.70}\\
    \bottomrule
    \end{tabular}
    }
    \caption{Controllability comparison on KITTI~\cite{Geiger2012CVPR} \textit{validation} set, showcasing vehicle mAP obtained by MonoFlex~\cite{zhang2021objects} using data generated by PerLDiff and the baseline BEVControl*. “NuScenes $\rightarrow$ KITTI” denotes initial training on NuScenes \textit{train} set followed by fine-tuning on KITTI \textit{train} set. \textbf{Bold} indicates the best result.}
    \label{Tab.KITTI}
     \vspace{-1em}
\end{table}

\begin{table*}[t]
    \centering
    % \vspace{-1em}
    \resizebox{0.95\textwidth}{!}{
    \begin{tabular}{@{}l|c|ccccc@{}}
    \toprule
    {Training} & Detector & mAP$\uparrow$ & NDS$\uparrow$ & mATE$\downarrow$ & mASE$\downarrow$ & mAOE$\downarrow$   \\ \midrule
    \textit{train}  & \multirow{4}{*}{BEVFormer} & 28.97 & 42.52 & 72.90 & 28.15 & 56.34 \\
    \textit{train} + Real \textit{val} &   & 32.20 & 45.44  & 69.43 & 27.40 & 52.88\\
    \textit{train} + Syn. \textit{val}* &  & 29.92 \textcolor{orange}{(-2.28\%)}& 43.20 \textcolor{orange}{(--2.24\%)}& 70.76 \textcolor{orange}{(+1.33\%)}& 27.69 \textcolor{orange}{(+0.29\%)}& 57.57 \textcolor{orange}{(+4.69\%)}\\
    \textit{train} + Syn. \textit{val} (Ours) &  &  \textbf{31.66} \textcolor{orange}{(-0.54\%)}&  \textbf{44.91} \textcolor{orange}{(-0.53\%)}&  \textbf{70.09} \textcolor{orange}{(+0.66\%)}&  \textbf{27.56} \textcolor{orange}{(+0.16\%)}&  \textbf{55.05}  \textcolor{orange}{(+2.17\%)}\\
    \midrule
    \textit{train}  & \multirow{4}{*}{StreamPETR} & 47.84 & 56.66 & 55.91 & 25.81 & 47.40 \\
    \textit{train} + Real \textit{val} &  & 50.92 & 58.68  & 54.36 & 25.12 & 45.36 \\
    \textit{train} + Syn. \textit{val}* &  & 47.37 \textcolor{orange}{(-3.55\%)} & 56.40 \textcolor{orange}{(-2.28\%)} & 56.99 \textcolor{orange}{(+2.63\%)} & 25.58 \textcolor{orange}{(+0.46\%)} & 47.69 \textcolor{orange}{(+2.33\%)}  \\
    \textit{train} + Syn. \textit{val} (Ours) & &  \textbf{49.07} \textcolor{orange}{(-1.85\%)} &  \textbf{57.92} \textcolor{orange}{(-0.76\%)} &  \textbf{55.71} \textcolor{orange}{(+1.35\%)} &  \textbf{25.57} \textcolor{orange}{(+0.45\%)} &  \textbf{47.08} \textcolor{orange}{(+1.72\%)} \\
    \bottomrule
    \end{tabular}
    }
    \caption{Performance comparison for the boosting performance of 3D detection models using synthesized dataset on NuScenes~\cite{caesar2020nuscenes} \textit{test} set using BEVFormer~\cite{li2022bevformer} and StreamPETR~\cite{wang2023exploring}. The “\textit{train} + Real \textit{val}” configuration serves as a benchmark, representing the ideal upper performance limit achievable. “Syn. \textit{val}*” represents the synthetic \textit{validation} set generated by BEVControl. The \textcolor{orange}{numbers} in parentheses indicate the performance disparity relative to the “\textit{train} + Real \textit{val}” configuration. \textbf{Bold} indicates the best result.}
    \label{Tab.augmentation}
     \vspace{-1em}
\end{table*}

The superiority of PerLDiff is further affirmed by BEV segmentation metrics~\cite{zhou2022cross} presented in Tab.~\ref{table:cvt_seg}. It significantly outperforms BEVControl* with a 0.52\% increase in Road mIoU and a 4.66\% increase in Vehicle mIoU, validating the efficacy of the PerL-based cross-attention mechanism in enhancing scene controllability. 
Regarding the Fréchet Inception Distance (FID)~\cite{heusel2017gans} metric, our results are comparable to those of BEVControl*. While PerLDiff incorporates prior constraints to ensure accuracy in object detection, this may adversely affect the details in the background of the images. As illustrated in the Appendix, PerLDiff produces background details that do not align with those of real images.

Compared to the state-of-the-art MagicDrive~\cite{gao2023magicdrive}, our method demonstrates superior performance across all metrics, particularly in the FID metric, reflecting an improvement of 2.84\%. Additionally, we achieve a 2.94\% improvement in mAP and a 0.73\% increase in NDS. These results substantiate the strengths of PerLDiff in terms of both generation quality and controllability.

%
% \noindent\textbf{Controllable Generation on KITTI.} 
\subsubsection{Controllable Generation on KITTI}
\begin{figure}[t]
    \centering
    \includegraphics[width=0.9\linewidth, keepaspectratio]{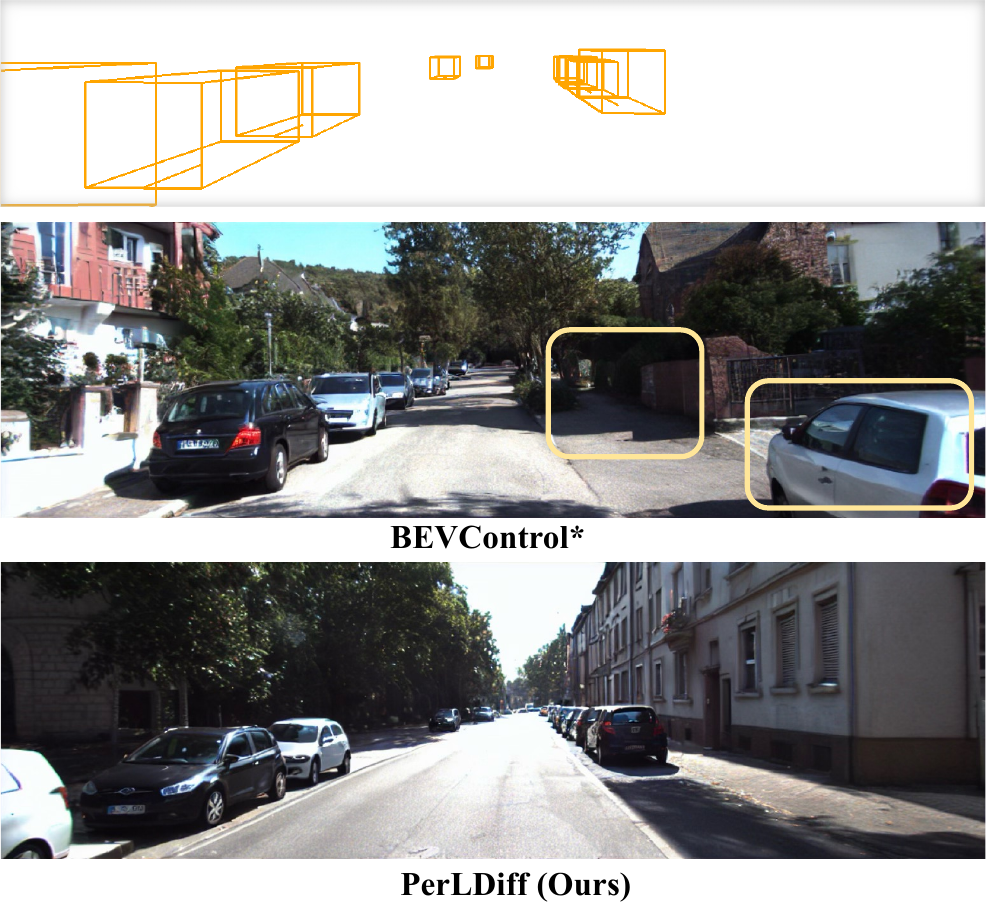}
    \vspace{-1em}
    \caption{Qualitative visualization comparison on KITTI~\cite{Geiger2012CVPR} dataset. \textcolor{darkyellow}{Yellow} markers denote instances where BEVControl* inaccurately generates output compared to PerLDiff.}
    \label{fig:kitti_img_1}
    \vspace{-1em}
\end{figure}

The scarcity of training data in the KITTI dataset~\cite{Geiger2012CVPR} often limits a generative model's ability to understand the relationship between control information and image synthesis. To address this challenge, we implement two distinct strategies for generating images within the KITTI framework: one strategy involves direct training using the KITTI \textit{train} set, while the other entails initial training on the NuScenes \textit{train} set followed by fine-tuning on the KITTI \textit{train} set. In Tab.~\ref{Tab.KITTI}, we present the results of monocular 3D object detection on the KITTI \textit{validation} set, utilizing a pretrained MonoFlex~\cite{zhang2021objects} detector. 
As illustrated in Fig.~\ref{fig:kitti_img_1}, the naive approach results in significant misalignment between the labels and the corresponding generated images, leading to a considerable performance gap: 11.04 vs. 0.33 for one metric and 13.12 vs. 1.32 for another. 
There are two main reasons for the observed differences. First, the limited size of the KITTI dataset, which contains just 3,712 training images, impedes the learning process of traditional methods that do not utilize PerL masking map. Second, monocular 3D object detection is highly sensitive to accurate depth prediction. Depth is derived from the 2D projected size and the estimated 3D size through perspective projection. Our method produces more precise object sizing, thereby enhancing detection performance. More visual results on KITTI can be found in the Appendix.

\begin{table*}[t]
    \centering
    % \vspace{-1em}
    \resizebox{0.95\textwidth}{!}{
    \begin{tabular}{@{}ccccccc|cc@{}}
    \toprule
    \multirow{2}{*}{Method} & Road & Box & \multirow{2}{*}{FID$\downarrow$} & \multirow{2}{*}{mAP$\uparrow$} & \multirow{2}{*}{NDS$\uparrow$} & \multirow{2}{*}{mAOE$\downarrow$} & Road & Vehicle \\
    &Mask&Mask&&&&&mIoU$\uparrow$&mIoU$\uparrow$\\ 
    \midrule
    Oracle &  &  & - & 27.06 & 41.89 & 0.54 &70.35 & 33.36 \\
    \midrule
    (a) &  &  & \textbf{13.05} &  16.48 & 28.08 & 0.88 &  60.74 & 22.47 \\
    (b) & \checkmark & & 13.20 & 16.27 & 28.40 & 0.86 & 61.18 & 23.14\\
    (c) &  & \checkmark &  13.54  & \textbf{26.07} & 36.07 & 0.74 & 61.21 & 26.27 \\
    (d) & \checkmark & \checkmark &  13.36 & 25.10 \textcolor{orange}{(+8.62\%)}& \textbf{36.24} \textcolor{orange}{(+8.16\%)} & \textbf{0.72} \textcolor{orange}{(-0.16\%)}& \textbf{61.26} \textcolor{orange}{(+0.52\%)}& \textbf{27.13} \textcolor{orange}{(+4.66\%)}\\
    \bottomrule
    \end{tabular}
    }
    \caption{Ablation of the PerL-based cross-attention, reporting 3D object detection improvements using BEVFormer and BEV segmentation enhancements using CVT. \textcolor{orange}{Numbers} in parentheses indicate performance gains over the baseline. \textbf{Bold} indicates the best result.}
    \label{table:prior_ablation}
\end{table*}
\subsubsection{Boosting Perception Models Using Synthesized Dataset}
% \noindent\textbf{Boosting Perception Models Using Synthesized Dataset.} 
Generative models have become widely acknowledged as effective tools for data augmentation, thereby improving the generalization capabilities of perception models. To evaluate this, we leverage synthesized image dataset to improve the performance of various detection models on NuScenes \textit{test} set.
The gains presented in the second row of Tab.~\ref{Tab.augmentation} confirm that augmenting with data annotated optimally (\textit{i.e.}, using the combined real NuScenes \textit{train} + Real \textit{val} set) is beneficial. 
The performance of BEVFormer~\cite{li2022bevformer} and StreamPETR~\cite{wang2023exploring} improved significantly after the dataset was augmented with real \textit{validation} set. The most notable gains for BEVFormer were observed in the mAP and NDS metrics, which increased by 3.23\% and 2.92\%, respectively. Similarly, for StreamPETR, increases in mAP and NDS were recorded at 3.08\% and 2.02\%, respectively.

Furthermore, augmentations using synthetic \textit{validation} set yielded competitive improvements that almost matched the performance gains observed with real \textit{validation} set.
The gaps in performance metrics, such as mAP, NDS, and mAOE, were minimal, thus solidifying the value of synthetic augmentation compared to the \textit{train-only} baseline. 
Specifically, BEVFormer and StreamPETR exhibited only slight gaps in mAP (0.54\% and 1.85\%), NDS (0.53\% and 0.76\%) and mAOE (2.17\% and 1.72\%), respectively. In addition, these discrepancies were even less pronounced compared to BEVControl*, highlighting the effectiveness of the PerL-based cross-attention mechanism.

\begin{table}[t]
    \centering
    \resizebox{\linewidth}{!}{
        \begin{tabular}{@{}cc|cccc|cc@{}}
        \toprule
        \multirow{2}{*}{Method} & \multirow{2}{*}{$\lambda_s$ $\lambda_b$} & \multirow{2}{*}{FID$\downarrow$} & \multirow{2}{*}{mAP$\uparrow$} & \multirow{2}{*}{NDS$\uparrow$} & \multirow{2}{*}{mAOE$\downarrow$} & Road & Vehicle\\
         &&&&&&mIoU$\uparrow$&mIoU$\uparrow$\\
         \midrule
        Oracle & -- & -- & 27.06 & 41.89 & 0.54  &  70.35 & 33.36\\ 
        \midrule
        (a) & 1.0 1.0 & \textbf{12.87}  & 22.30 & 34.08 & 0.73 & 61.31 & 25.03 \\
        (b) & 3.0 3.0 & 14.03 & 24.41 & 35.75 & 0.74 & 60.58 & 26.82 \\
        \rowcolor{gray!25}(c)  & 5.0 5.0 & 13.36  & \textbf{25.10} & \textbf{36.24} & \textbf{0.72} & 61.26 & \textbf{27.13} \\
        (d) & 10.0 10.0 & 14.24 & 24.98 & 35.52 & 0.76 & \textbf{61.75} & 26.62 \\
        \bottomrule
        \end{tabular}
    }
    \caption{Ablation of different values of masking map weight coefficients $\lambda_s$ and $\lambda_b$ in PerL-based cross-attention. We report the 3D object detection results based on BEVFormer and BEV Segmentation results based on CVT. \textbf{Bold} indicates the best result.}
    \label{table:weight_bev_detection}
    \vspace{-1em}
\end{table}
\subsection{Ablation Study}
\label{sec:ablation_study}
To determine the effectiveness of the key components within PerLDiff, we perform ablation studies concentrated on key elements: PerL-based cross-attention. Additionally, we provide more qualitative results in the Appendix to further demonstrate the efficacy of PerLDiff compared to the baseline approaches, BEVControl* and MagicDrive.

% \noindent\textbf{Effectiveness of PerL-based Cross-attention.} 
\subsubsection{Effectiveness of PerL-based Cross-attention}
To illustrate the impact of PerL-based cross-attention, we devised a comprehensive comparative experiment, the results of which are presented in Tab.~\ref{table:prior_ablation}. Method (a) employs road map and 3D box as conditions, which are integrated into the model using standard cross-attention~\cite{yang2023bevcontrol,gao2023magicdrive} with the configuration mirroring that of BEVControl*. “Box Mask” and “Road Mask” denote the process wherein the control information is merged with the model through PerL-based cross-attention.
Method (a)$\rightarrow$(b) signifies the adoption of PerL-based cross-attention for the road map, leading to improvements of 0.44\% in Road mIoU and 0.32\% in NDS. These gains underscore the augmented controllability achieved by combining the road map with PerL-based cross-attention and its efficacy in aligning generated data with real observations. 
Additionally, Method (a)$\rightarrow$(c) results in marked improvements of 9.59\% in mAP, 7.99\% in NDS, 0.14\% in mAOE and 3.80\% in Vehicle mIoU, strongly supporting the utility of PerL-based cross-attention in producing accurate data-annotation alignments for objects. 
To optimally regulate elements of the background and foreground, Method (c), in contrast to baseline Method (a), indicates increases of 8.62\% in mAP, 8.16\% in NDS, 0.16\% in mAOE, 0.52\% in Road mIoU and 4.66\% in Vehicle mIoU. These results validate the efficiency of PerL-based cross-attention in enhancing image controllability.

% \noindent\textbf{Effectiveness of Masking Map Weight Coefficients.} 
\subsubsection{Effectiveness of Masking Map Weight Coefficients}
Tab.~\ref{table:weight_bev_detection} examines the effects of varying the masking map weight coefficients $\lambda_s$ and $\lambda_b$, where higher values indicate a greater integration of PerL knowledge into network learning. The table demonstrates that detection metrics improve with increasing values of $\lambda_s$ and $\lambda_b$ within a certain range. However, the FID score also increases, underscoring the significant role of PerL knowledge in the controllable learning process.
% of the diffusion model. 
For optimal controllability, we set the default values of $\lambda_s$ and $\lambda_b$ to 5.0 in the main text.

\subsubsection{Effectiveness of Different Road Map Encoders}
\label{subsubsec:road_map_enc}
% \textcolor{blue}{
In Tab.~\ref{table:effectiveness_roadmap_enc}, we present an ablation study comparing different road map encoders, replacing ConvNext~\cite{liu2022convnet} with alternatives such as the CLIP image encoder. ConvNext was initially chosen for its ability to effectively extract image features while maintaining a relatively low parameter count (27.82M). Additionally, it has demonstrated strong performance in downstream segmentation tasks, particularly in capturing edge information~\cite{liu2022convnet}.
As shown in Tab.~\ref{table:effectiveness_roadmap_enc}, the results indicate that the choice of feature extraction network has minimal impact on PerLDiff, with only a $\pm0.02$ NDS difference between the CLIP and ConvNext encoders. Therefore, we adopt ConvNext as the road map encoder and freeze its weights in the main results.
\begin{table}[t]
    \centering
    \resizebox{\linewidth}{!}{
    \begin{tabular}{@{}lc|ccccc@{}}
    \toprule
    Method & Detector & Parameter (M) & FID$\downarrow$ & mAP$\uparrow$ & NDS$\uparrow$ & mAOE$\downarrow$ \\
    \midrule
    CLIP encoder & BEVFormer & 303.97 & \textbf{12.77} & \textbf{25.41} & \textbf{36.26} & 0.73 \\
    ConvNext encoder & &  \textbf{27.82} & 13.36 & 25.10 & 36.24 & \textbf{0.72} \\
    \midrule
    CLIP encoder & BEVFusion & 303.97 & \textbf{12.77} & \textbf{15.80} & \textbf{24.45} & 0.81 \\
    ConvNext encoder& & 27.82 & 13.36 & 15.24 & 24.05 & \textbf{0.78} \\
    \bottomrule
    \end{tabular}
    }
    \caption{Ablation study comparing different road map encoders with frozen weights. We present 3D object detection results based on BEVFormer and BEVFusion. \textbf{Bold} indicates the best result. 
    }
    \label{table:effectiveness_roadmap_enc}
     % \vspace{-1em}
\end{table}

\section{Conclusion}
\label{sec:conclusion}
In conclusion, PerLDiff introduces a streamlined framework that adeptly merges geometric constraints with synthetic street view image generation, harnessing diffusion models' power for high-fidelity visuals. 
The architecture boasts a PerL-based controlling module (PerL-CM), through training, becomes seamlessly integrated with Stable Diffusion. Meanwhile, a cutting-edge PerL-based cross-attention mechanism guarantees meticulous feature guidance at the object level for precise control.
Experiments on NuScenes and KITTI datasets confirm PerLDiff's enhanced performance in image synthesis and downstream tasks like 3D object detection and segmentation. 
Flexible yet precise, PerLDiff's method of PerL-based cross-attention with geometric perspective projections during training finely balances image realism with accurate condition alignment.

\section*{Acknowledgement}
\label{sec:acknowledgment}
This work was supported by National Natural Science Foundation of China (No. 62476051, No. 62176047) and Sichuan Natural Science Foundation (No. 2024NSFTD0041).

% \clearpage
{
    \small
    \bibliographystyle{ieeenat_fullname}
    \bibliography{perldiff}
}

\clearpage
\setcounter{section}{0}
\renewcommand{\thesection}{\Alph{section}}
\maketitlesupplementary

The supplementary material is organized into the following sections:
\begin{itemize}
    \item Section~\ref{sec:preliminary_ddpm}: DDPM Preliminaries
    \item Section~\ref{sec:implementation_details}: Implementation Details
    \item Section~\ref{sec:limit_and_future}: Limitation and Future Work
    \item Section~\ref{sec:ablation_studies}: Additional Experiments
    \item Section~\ref{sec:visualization_results}: Visualization Results
\end{itemize}

\section{DDPM Preliminaries}
\label{sec:preliminary_ddpm}
Denoising Diffusion Probabilistic Models (DDPM)~\cite{ho2020denoising} are a class of generation models which simulate a Markov chain of diffusion steps to gradually convert data samples into pure noise. The generative process is then reversed to synthesize new samples from random noise. We commence with an observation \( x_0 \) sampled from the data's true distribution \( q(x) \), and then progressively apply Gaussian noise over a series of \( T \) time steps. The forward diffusion is mathematically defined as $q(x_t | x_{t-1}) = \mathcal{N}(x_t; \sqrt{1 - \beta_t} x_{t-1}, \beta_t \mathbf{I}), \label{eq:forward_diffusion}$, where \( \beta_t \) is a variance term that can be either time-dependent or learned during training. The entire forward diffusion process can be represented as the product of the conditional distributions from each step:
\begin{equation}
    q(x_{1:T} | x_0) = \prod_{t=1}^{T} q(x_t | x_{t-1}), \label{eq:full_forward_diffusion}
\end{equation}
where the sequence \( \{ \beta_t \}_{t=1}^{T} \) specifies the noise schedule applied at each timestep. The diffusion process is notable for permitting direct sampling of \( x_t \) from \( x_0 \) using a closed-form expression:
\begin{equation}
    q(x_t | x_0) = \sqrt{\bar{\alpha}_t} x_0 + \sqrt{1 - \bar{\alpha}_t} \epsilon, \quad \text{where} \quad \epsilon \sim \mathcal{N}(0, \mathbf{I}), \label{eq:direct_sampling}
\end{equation}
in which \( \alpha_t = 1 - \beta_t \) and the cumulative product \( \bar{\alpha}_t = \prod_{s=1}^{t} \alpha_s \). To synthesize new samples, a reverse process known as the backward diffusion is learned, which conceptually undoes the forward diffusion. This inverse transition is captured through a parameterized Gaussian distribution:
\begin{equation}
    p_{\theta}(x_{t-1} | x_t) = \mathcal{N}(x_{t-1}; \mu_{\theta}(x_t), \sigma_{\theta}^2(x_t) \mathbf{I}). \label{eq:backward_diffusion}
\end{equation}
\begin{figure}[t]
    \centering
    \includegraphics[width=\linewidth, keepaspectratio]{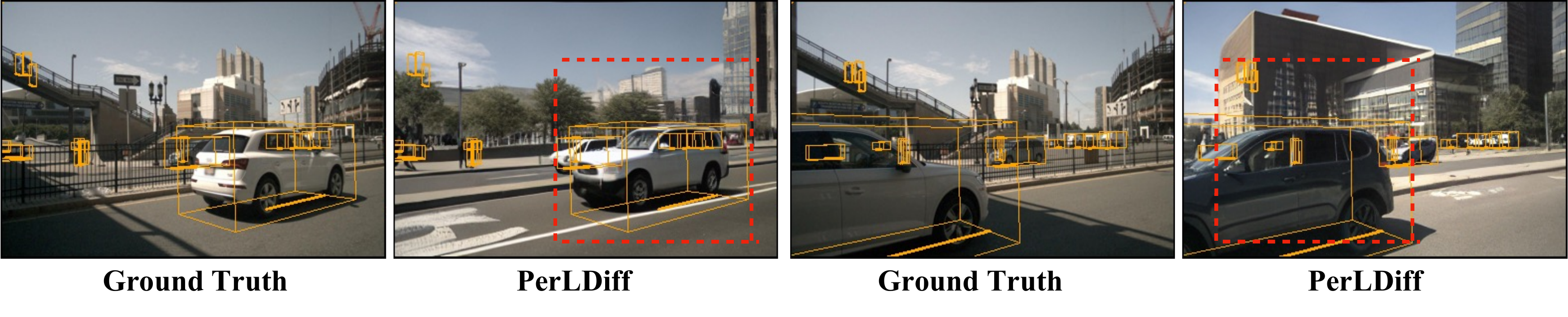}
    \vspace{-1.5em}
    \caption{Failure cases of PerLDiff, with \textcolor{red}{red} markers highlighting instances where, compared to the ground truth, PerLDiff generates images with the front and rear of vehicles reversed.}
    \label{fig:PerLdiff_limit}
    \vspace{-1em}
\end{figure}
\section{Implementation Details}
\label{sec:implementation_details}
PerLDiff utilizes the pre-trained Stable Diffusion v1.4~\cite{rombach2022high}, augmented with specific modifications to enhance scene control. Training was conducted on a server equipped with eight Tesla V100 (32 GB) GPUs over 60,000 iterations, which required two days.
An initial batch size of 16 was adjusted to a per-GPU batch of two for focused optimization, particularly for data samples comprising six view images per frame. The generation of samples conforms to the CFG rule~\cite{ho2022classifier}, employing a guidance scale of 5.0 and the DDIM~\cite{song2020denoising} across 50 steps. 

For scene manipulation, the text encoder within Stable Diffusion is retained, along with a weight-frozen CLIP to manage textual inputs and $\text{ConvNext}$ for processing road maps. Feature extraction from PerL boxes is conducted via an MLP, optimized through PerL-based controlling module (PerL-CM) with randomly initialized weights. In contrast, certain modules inherit and freeze pre-trained weights from Stable Diffusion. The key parameters within PerL-CM, $\lambda_b$ and $\lambda_s$, are set to 5.0 to facilitate optimal image synthesis. Furthermore, DDIM~\cite{song2020denoising} and CFG~\cite{ho2022classifier} are integrated into our training regimen, with a novel approach of omitting all conditions at a rate 10\% to foster model versatility.

The optimization process employs AdamW~\cite{loshchilov2017decoupled} without a weight decay coefficient and with a learning rate of $5 \times 10^{-5}$, complemented by a warm-up strategy during the first 1,000 iterations. BEVFormer~\cite{li2022bevformer}, StreamPETR~\cite{wang2023exploring}, and CVT~\cite{zhou2022cross} were retrained using original configurations tailored to our target resolution. The performance of BEVFusion~\cite{liu2023bevfusion} and MonoFlex~\cite{zhang2021objects} was assessed using their provided code and pre-trained weights.
\section{Limitation and Future Work.}
\label{sec:limit_and_future}
Fig.~\ref{fig:PerLdiff_limit} depicts several failure cases of PerLDiff, where the model erroneously generates vehicles with the front and rear orientations reversed, in contrast to the ground truth. This limitation arises from the usage of a PerL mask in PerLDiff, which does not account for the orientation on the 2D PerL plane. Future endeavors may explore video generation, extending to work such as DrivingDiffusion~\cite{li2023drivingdiffusion}, Panacea~\cite{wen2023panacea}, and Driving into the Future~\cite{wang2023driving}.
\section{Additional Experiments}
\label{sec:ablation_studies}
In this section, we present additional experiments conducted to validate controllability at different resolutions (256$\times$ 704) and to assess the contributions of individual components within our PerLDiff. Our studies focus on the following aspects:
\begin{itemize}
    \item Effectiveness of Controllable Generation on NuScenes (Subsection~\ref{subsec:effectiveness_controllable_generation_nuscenes_2})
    \item Effectiveness of Perl-based Cross Attention (Object)
    (Subsection~\ref{subsec:detail_of_perl_based})
    \item Effectiveness of View Cross-attention for Multi-View Consistency (Subsection~\ref{subsec:effectiveness_view_cross_attention})
    \item Effectiveness of PerLDiff Based on ControlNet
    (Subsection~\ref{subsec:effectiveness_controlnet})
    \item Effectiveness of Classifier-Free Guidance Scale (Subsection~\ref{subsec:effectiveness_classifier_free_guidance_scale})

\end{itemize}
Our results confirm the superior performance of our method across various resolutions and illustrate how each component is integral to the success of our PerLDiff. 
\subsection{Effectiveness of Controllable Generation on NuScenes}
\label{subsec:effectiveness_controllable_generation_nuscenes_2}
In Tab.~\ref{table:effectiveness_controllable_generation_nuscenes_2}, we conduct a comparative analysis to emphasize the capabilities of PerLDiff for controllable generation at a resolution of 256$\times$704. This quantitative evaluation contrasts our method with other leading approaches based on the detection metrics provided by BEVFusion~\cite{liu2023bevfusion}. Our PerLDiff exhibits significantly superior performance, achieving mAP improvements of 3.84\% and 11.50\%, and NDS increases of 0.45\% and 10.80\%, compared to MagicDrive~\cite{gao2023magicdrive} and BEVControl*, respectively. These results confirm the efficacy of PerLDiff in the precise controllable generation at the object level.
\begin{table}[t]
    \centering
    \caption{Controllability comparison for street view image generation on the NuScenes \textit{validation} set. A quantitative evaluation using 3D object detection metrics from BEVFusion~\cite{liu2023bevfusion}.}
    \label{table:effectiveness_controllable_generation_nuscenes_2}
    % \vspace{-1em}
    \resizebox{\linewidth}{!}{
    \begin{tabular}{@{}lc|ccccc@{}}
    \toprule
    Method  & FID$\downarrow$ & mAP$\uparrow$ & NDS$\uparrow$ &  mATE$\downarrow$&  mASE$\downarrow$&mAOE$\downarrow$\\
    \midrule
    Oracle & -- & 35.54 & 41.20 & 0.67 & 0.27 & 0.56  \\
    \midrule
    MagicDrive~\cite{gao2023magicdrive} & 16.59  & 20.85 & 30.26 & -- & -- & -- \\
    BEVControl* & 15.94  & 13.19 & 19.91 & 0.94 & 0.34 & 0.96  \\
    \rowcolor{gray!25}PerLDiff (Ours) & \textbf{15.67}  &\textbf{24.69} & \textbf{30.71} &\textbf{0.82} &  \textbf{0.28} & \textbf{0.76}\\
    \bottomrule
    \end{tabular}
    }
\end{table}

\subsection{Effectiveness of Perl-based Cross Attention (Object)}
\label{subsec:detail_of_perl_based}
\begin{figure}[t]
  \centering
  \includegraphics[width=\linewidth, keepaspectratio]{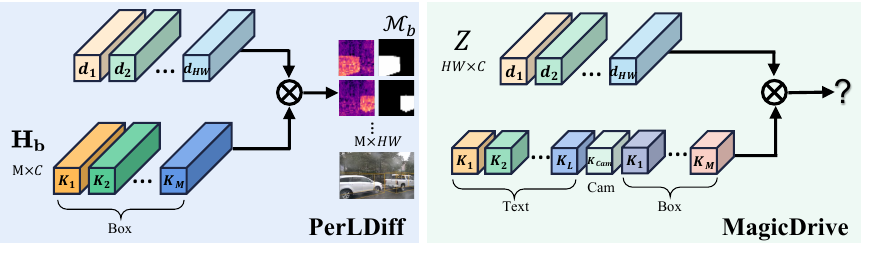}
  \caption{Overview of the PerL-based cross-attention (Object). MagicDrive employs text cross-attention to create a unified feature, while PerLDiff uses the PerL masking map to allow for precise control of pixel features for each object.}
  \label{fig:magic_mask}
\end{figure}

To facilitate a better understanding of PerLDiff, we provide a detailed explanation of the PerL-based cross-attention (Object). As shown in Fig.~\ref{fig:magic_mask}, MagicDrive~\cite{gao2023magicdrive} utilizes text cross-attention from Stable Diffusion to implicitly learn a unified feature that concatenates text, camera parameters, and bounding boxes in the token dimension. In contrast, PerLDiff employs the PerL masking map as a prior, allowing each object condition to precisely control the corresponding pixel features. This results in more accurate positioning and orientation of objects in the generated images. Additionally, we integrated the object mask into the token dimension corresponding to the bounding box. As shown in Tab.~\ref{table:magic_mask}, the results indicate improvements in BEVFormer, with NDS (e.g., 29.77 vs. 28.79 for MagicDrive) and mAOE (e.g., 0.73 vs. 0.81 for MagicDrive) demonstrating the effectiveness of PerLDiff in enhancing the performance of MagicDrive. Note that MagicDrive utilizes a single attention map for managing text, camera parameters, and boxes in the cross-attention process. Consequently, our ability to make improvements is constrained by the limited scope available for modifying the attention map within this architecture.
% \vspace{-1.5em}
%
\begin{table}[t]
    \centering
    \caption{Impact of integrating the PerL masking map (Object) into MagicDrive. We present the 3D object detection results based on BEVFormer~\cite{li2022bevformer}, with outcomes showing superior performance emphasized in \textbf{bold}.}
    \label{table:magic_mask}
    \resizebox{\linewidth}{!}{
    \begin{tabular}{@{}l|cccccc@{}}
    \toprule
    Method & FID$\downarrow$ & mAP$\uparrow$ & NDS$\uparrow$ & mAOE$\downarrow$ & mAVE$\downarrow$ & mATE$\downarrow$\\
    \midrule
    MagicDrive & \textbf{15.92} & 15.21 & 28.79 & 0.81  &  0.57  & 0.95 \\
    MagicDrive + Mask & 16.68 & \textbf{15.54} & \textbf{29.77} & \textbf{0.73} & \textbf{0.56} & \textbf{0.89}  \\
    \bottomrule
    \end{tabular}
    }
\end{table}

\subsection{Effectiveness of View Cross-attention for Multi-View Consistency}  
\label{subsec:effectiveness_view_cross_attention}
\begin{figure}[t]
    \centering
    \includegraphics[width=\linewidth]{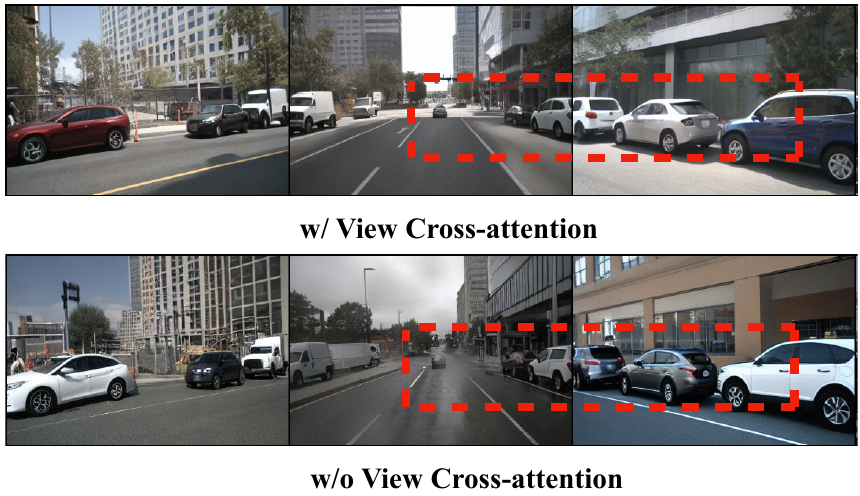}
    \vspace{-1.5em}
    \caption{Comparative visualization of outputs with (Top) and without (Bottom) view cross-attention. \textcolor{red}{Red} markers highlight discontinuities in the images generated without view cross-attention.}
    \label{fig:effectiveness_view_cross_attention}
\end{figure}

View cross-attention ensures the seamless integration of visual data by maintaining continuity and consistency across the multiple camera feeds that are integral to current multi-functional perception systems in autonomous vehicles. Typically, autonomous vehicles feature a 360-degree horizontal surround view from a BEV perspective, resulting in overlapping fields of vision between adjacent cameras. Consequently, we facilitate direct interaction between the noise maps of each camera and those of the immediate left and right cameras. Given the noisy images from the current, left, and right cameras, designated as $\mathbf{Z}_{b}$, $\mathbf{Z}_{l}$, and $\mathbf{Z}_{r}$, respectively, the output of this multi-view generation is given by:
\begin{align}\label{eq:perspective interaction}
\mathbf{\hat{Z}} = \mathbf{Z}_{b} + \mathcal{C}(\mathbf{Z}_{b}, \mathbf{Z}_{l}, \mathbf{Z}_{l}) + \mathcal{C}(\mathbf{Z}_{b}, \mathbf{Z}_{r}, \mathbf{Z}_{r}),
\end{align}
where $\mathcal{C}(\cdot)$ represents the standard cross-attention operation, which accepts three input parameters: query, key, and value, respectively. This approach systematically integrates spatial information from various viewpoints, enabling the synthesis of images that exhibit visual consistency across distinct camera perspectives.
Fig.\ref{fig:effectiveness_view_cross_attention} offers a visual comparison of the model output with and without the application of view cross-attention. 
Upon integrating view cross-attention into PerLDiff, the procedure of the PerL-CM is detailed in Algo.~\ref{alg:PerL_cm}.

\begin{algorithm}[t]
\caption{PerL-based Controlling Module (PerL-CM)}
\label{alg:PerL_cm}
\begin{algorithmic}[1]
\Statex \textbf{Input:} road map features $\mathbf{H}_m \in \mathbb{R}^{1 \times C}$, road masking map $\mathcal{M}_s \in \mathbb{R}^{HW \times 1}$, box features $\mathbf{H}_b \in \mathbb{R}^{M \times C}$, box masking map $\mathcal{M}_b \in \mathbb{R}^{HW \times M}$, scene text description features $\mathbf{H}_d \in \mathbb{R}^{1 \times C}$, noisy multi-view street image feature $\mathbf{Z} \in \mathbb{R}^{HW \times C}$, and dimension $d$ (omit the detail of multi-view perspectives)
\Statex \textbf{Output:} Updated $\mathbf{Z}$
\State $\mathcal{A}_s \gets \textit{softmax}(\lambda_s \cdot \mathcal{M}_s + \mathbf{Z}\mathbf{H}_m^{T}/\sqrt{d})$
\Statex \textit{// compute attention map for the road map in PerL-based cross-attention (scene)}
\State $\mathbf{Z_s} \gets \gamma_s \cdot\mathcal{A}_s \mathbf{H}_m + \mathbf{Z}$ 
\State $\mathcal{A}_b \gets \textit{softmax}(\lambda_b \cdot \mathcal{M}_b + \mathbf{Z_s}\mathbf{H}_b^{T}/\sqrt{d})$
\Statex \textit{// compute attention map for the box in PerL-based cross-attention (object)} 
\State $\mathbf{Z_b} \gets \gamma_b\cdot \mathcal{A}_b \mathbf{H}_b + \mathbf{Z_s}$
\State $\mathbf{\hat{Z}} \gets \mathbf{Z}_{b} + \mathcal{C}(\mathbf{Z}_{b}, \mathbf{Z}_{l}, \mathbf{Z}_{l}) + \mathcal{C}(\mathbf{Z}_{b}, \mathbf{Z}_{r}, \mathbf{Z}_{r})$ 
\Statex \textit{// maintain visual consistency via View cross-attention}
\State $\mathbf{Z^{*}} \gets \textit{softmax}(\mathbf{\hat{Z}}\mathbf{H}_d^{T}/\sqrt{d})\mathbf{H}_d + \mathbf{\hat{Z}}$ 
\Statex \textit{// alter illumination and atmospheric effects by Text cross-attention}
\end{algorithmic}
\end{algorithm}
\begin{table}[t]
    \centering
    \caption{Ablation study comparing PerLDiff with a ControlNet-based model. We present 3D object detection results based on BEVFormer, BEVFusion. Outcomes demonstrating superior performance are highlighted in \textbf{bold}.}
    \label{table:effectiveness_controlnet}
    \resizebox{\linewidth}{!}{
    \begin{tabular}{@{}lcc|ccc@{}}
    \toprule
    Method & Detector &FID$\downarrow$ & mAP$\uparrow$& NDS$\uparrow$ & mAOE$\downarrow$ \\
    \midrule
    ControlNet-based & BEVFormer & 20.46 & 18.07 & 28.48 & 0.87   \\
    GLIGEN-based &  & \textbf{13.36}  & \textbf{25.10} & \textbf{36.24} & \textbf{0.72} \\
    \midrule
    ControlNet-based & BEVFusion & 20.46 & 10.45 & 15.29 & 0.89   \\
    GLIGEN-based &  & \textbf{13.36}  & \textbf{15.24} & \textbf{24.05} & \textbf{0.78} \\
    \bottomrule
    \end{tabular}
    }
\end{table}
\begin{figure}[t]
  \centering
  \includegraphics[width=\linewidth, keepaspectratio]{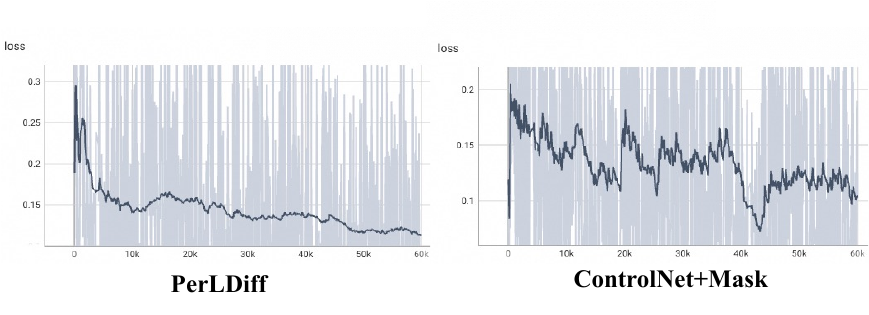}
  \caption{Training curves of PerLDiff and ControlNet-based, illustrating that Ours converges more rapidly during training.}
  \label{fig:effectiveness_controlnet}
\end{figure}
\begin{figure*}[t]
  \centering
    \vspace{-1.5em}
  \includegraphics[width=0.95\textwidth, keepaspectratio]{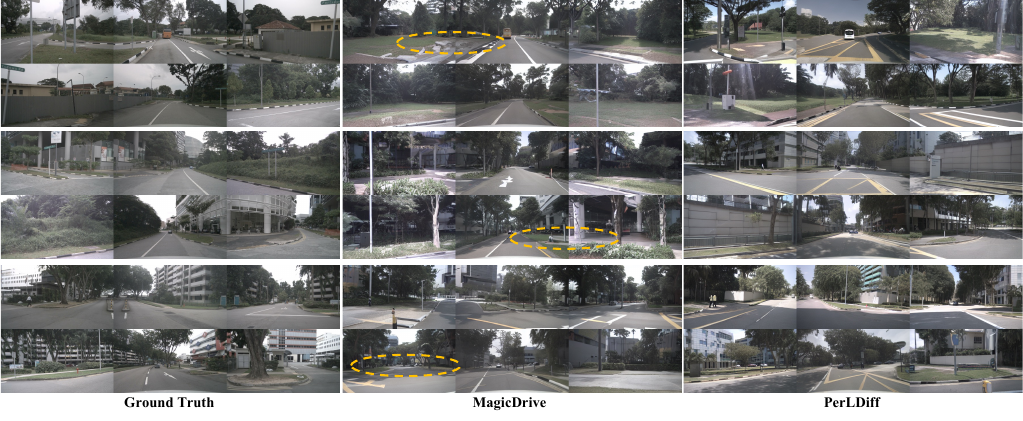}
  \caption{Qualitative comparison with MagicDrive. For \textbf{scene controllability}, PerLDiff demonstrates superior performance by results consistent with ground truth road information. Regions highlighted by \textcolor{yellow}{yellow} circles indicate areas where fail to align with ground truth.}
  \label{fig:scene_control}
  \vspace{-1em}
\end{figure*}
\subsection{Effectiveness of PerLDiff Based on ControlNet}
\label{subsec:effectiveness_controlnet}
In Tab.~\ref{table:effectiveness_controlnet}, we present an ablation study that replaces the architecture of PerLDiff with a ControlNet-based~\cite{zhang2023adding} model trained only on view cross-attention in Stable Diffusion. As shown in Tab.\ref{table:effectiveness_controlnet}, the performance of the ControlNet-based model is inferior to that of PerLDiff. Furthermore, Fig.~\ref{fig:effectiveness_controlnet} illustrates that PerLDiff employs a network architecture similar to GLIGEN~\cite{li2023gligen}, allowing it to converge more quickly on smaller datasets, such as NuScenes, compared to the ControlNet-based.
% \vspace{-1.5em}

\subsection{Effectiveness of Classifier-Free Guidance Scale}
\label{subsec:effectiveness_classifier_free_guidance_scale}
In Tab.~\ref{table:scale_bev_detection}, we assess the effect of the CFG~\cite{ho2022classifier} scale on the sampling of data generation. The term “scale” refers to the CFG scale, which is adjusted to balance conditional and unconditional generation. 
The transition from Method (b) to (e) indicates an increase in the CFG scale from 5.0 to 12.5. The results show an average increase of 2.87 in FID, an average decrease of 0.87\% in mAP, an average reduction of 1.03\% in NDS, a 0.02\% increase in mAOE and a 1.07\% drop in Vehicle mIoU. This provides substantial evidence that an excessively large CFG scale can degrade the quality of generated images and adversely affect various performance metrics.

\begin{table}[t]
    \centering
    \caption{Comparison of different CFG~\cite{ho2022classifier} scale to each metric. We report the 3D object detection results based on BEVFormer~\cite{li2022bevformer} and BEV Segmentation results based on CVT~\cite{zhou2022cross}.}
    \label{table:scale_bev_detection}
    % \vspace{-1em}
    \resizebox{\linewidth}{!}{
    \begin{tabular}{@{}lc|cccc|cc@{}}
    \toprule
    \multirow{2}{*}{Method} & \multirow{2}{*}{scale} & \multirow{2}{*}{FID$\downarrow$} & \multirow{2}{*}{mAP$\uparrow$}
     & \multirow{2}{*}{NDS$\uparrow$} & \multirow{2}{*}{mAOE$\downarrow$} & Road & Vehicle\\
    &&&&&&mIoU$\uparrow$&mIoU$\uparrow$\\
    \midrule
    Oracle & -- & -- & 27.06 & 41.89 & 0.54  &  70.35 & 33.36\\ 
    \midrule
    (a) & 2.5 & \textbf{12.36}  & 23.89 & 36.03 & \textbf{0.70} & 60.05 & 26.95 \\
    \rowcolor{gray!25}(b) & 5.0 & 13.36  & \textbf{25.10} & \textbf{36.24} & 0.72 & 61.26 & \textbf{27.13} \\
    (c) & 7.5 & 15.52 & 24.62 & 35.60 & 0.74 & \textbf{61.52} & 26.63 \\
    (d) & 10.0 & 16.32 & 24.20 & 35.05 & 0.73 & 61.43 & 26.00  \\
    (e) & 12.5 & 16.86 & 23.86 & 34.98 & 0.74 & 61.25 & 25.55\\
    \bottomrule
    \end{tabular}
    }
      \vspace{-1em}
\end{table}
\begin{figure}[t]
    \centering
    \includegraphics[width=\linewidth, keepaspectratio]{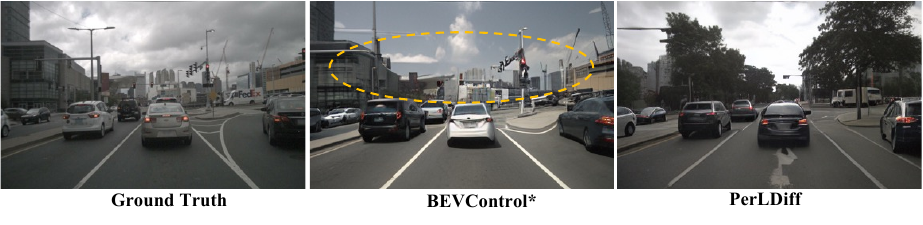}
    \caption{Qualitative results of the generated images reveal discrepancies in background details. As indicated by the \textcolor{yellow}{yellow} circle, PerLDiff produces background elements that do not align with real images due to the incorporation of the PerL masking map.}
    \label{fig:fid}
      \vspace{-1em}
\end{figure}
\section{Visualization Results}
\label{sec:visualization_results}

\begin{figure*}[t]
  \vspace{-1em}
  \includegraphics[width=\textwidth, keepaspectratio]{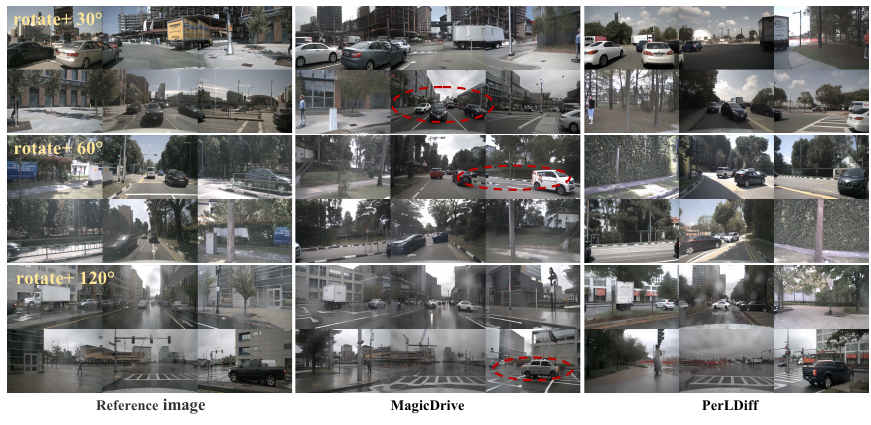}
  \caption{Qualitative comparison with MagicDrive. For \textbf{object controllability}, PerLDiff exhibits superior performance by generating objects at arbitrary angles. Regions highlighted by \textcolor{red}{red} circles denote scenarios where the images fail to achieve correct orientation.}
  \label{fig:object_control}
    \vspace{-1em}
\end{figure*}

\begin{figure*}[t]
    \centering
      \vspace{-1em}
    \includegraphics[width=\textwidth, keepaspectratio]{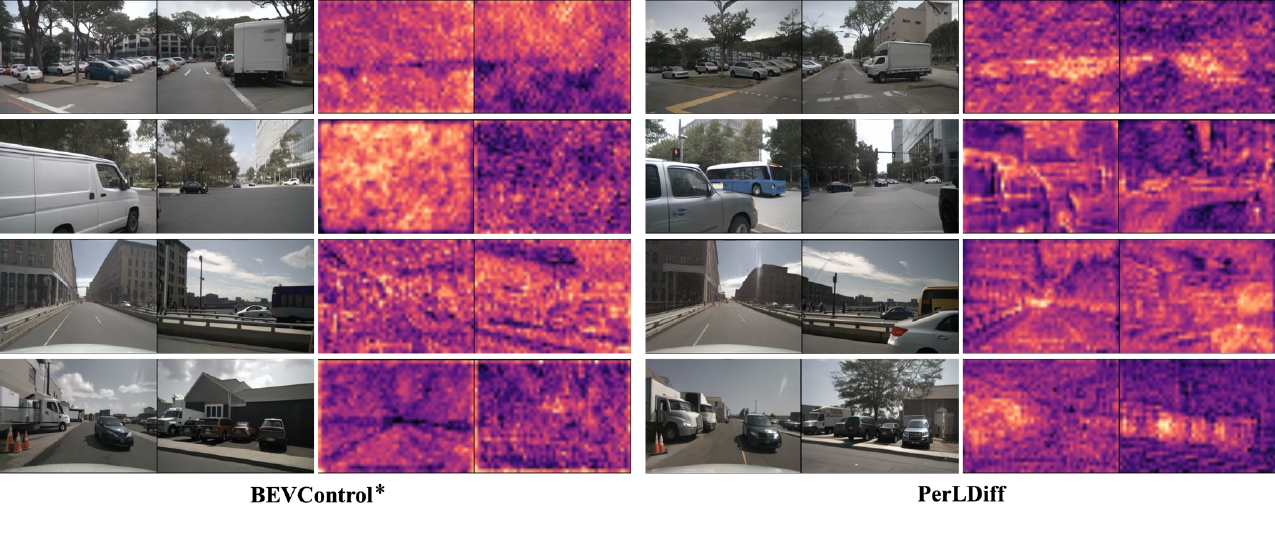}
    \caption{Visualization of cross-attention map results. From left to right, we present the generated images and corresponding cross-attention maps from our baseline BEVControl* and our PerLDiff. BEVControl* produces disorganized and vague attention maps, which result in inferior image quality. Conversely, PerLDiff method fine-tunes the response within the attention maps, resulting in more accurate control information at the object level and improved image quality.}
    \label{fig:PerLdiff_vs_base_atten_2}
      \vspace{-1em}
\end{figure*}

To further demonstrate the controllable generation capabilities of PerLDiff, we present additional visual results. 
Fig.~\ref{fig:scene_control} offers extended examples illustrating the superiority of PerLDiff in scene controllability, while Fig.~\ref{fig:object_control} highlights its effectiveness in controlling object orientation.
Fig.~\ref{fig:PerLdiff_vs_base_atten_2} reveals that BEVControl* produces chaotic and indistinct attention maps leading to suboptimal controllability, PerLDiff optimizes the response areas of the attention map, resulting in accurate object-level control.

\begin{table}[h!]
    \centering
    \vspace{-1em}
    \caption{Comparison with Panacea on synthetic 256$\times$704 validation data.
    The quantitative evaluation using 3D object detection metrics from StreamPETR~\cite{wang2023exploring}.
    }
    % \vspace{-1em}
    \label{tab:panacea}
    \resizebox{0.48\textwidth}{!}{%
        \begin{tabular}{l|c|ccccc}
        \toprule
        Method      & FID↓   & mAP↑   & NDS↑   & mAOE↓ & mATE↓ & mAVE↓ \\
        \midrule
        Oracle        &--     & 47.05  & 56.24  & 0.37   & 0.61  & 0.27    \\
        \midrule
        Penacea       & 16.96  & 22.50  & 36.10  & 0.73  & --    & 0.47  \\
        \textbf{PerLDiff}  & \textbf{15.67} & \textbf{35.09} & \textbf{44.19} & \textbf{0.64} & \textbf{0.75} & \textbf{0.45} \\
        \bottomrule
    \end{tabular}%
    % \vspace{-4em}
}
\end{table}

\begin{figure*}[t]
    \centering
      \vspace{-1em}
    \includegraphics[width=\textwidth, keepaspectratio]{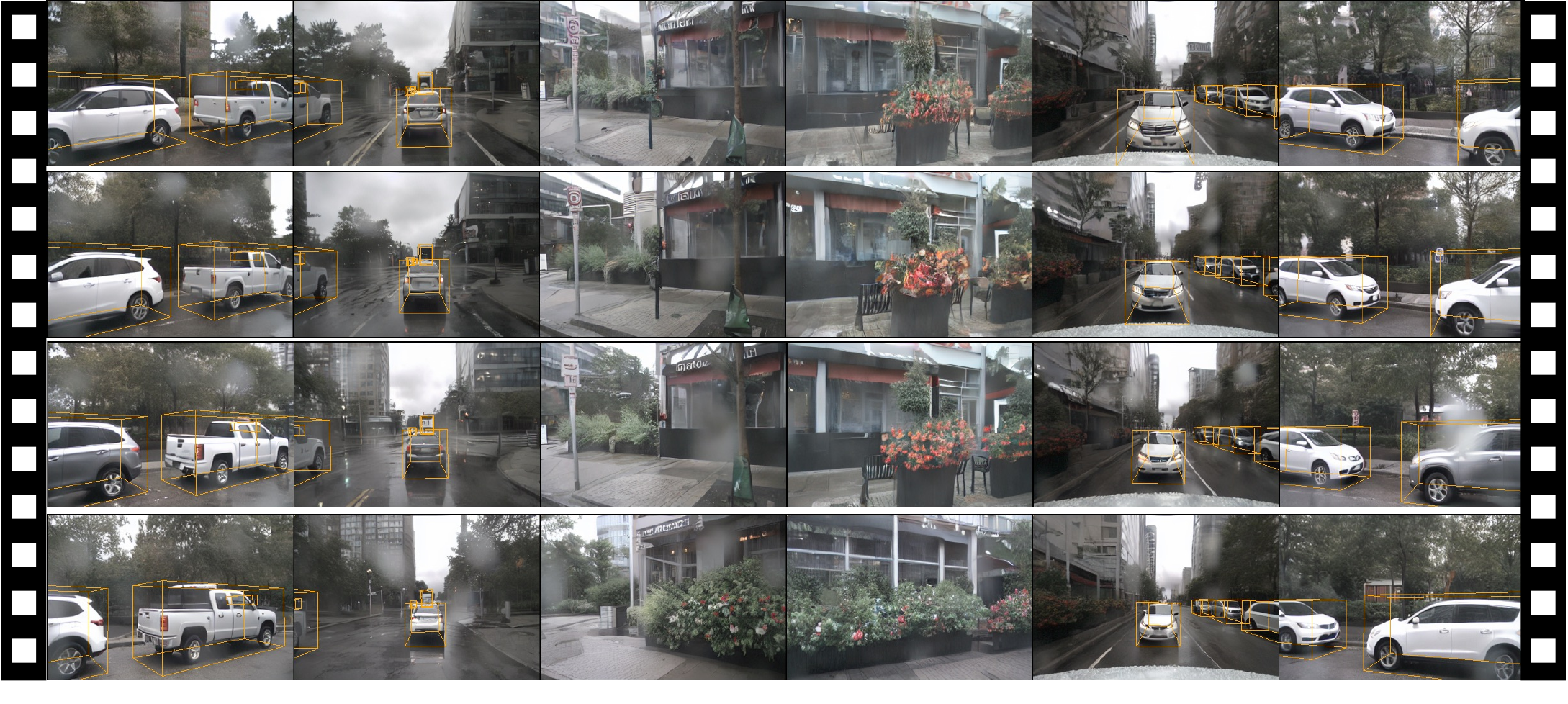}
    \caption{Qualitative visualizations from the NuScenes. PerLDiff demonstrate consistent alignment of object positions and categories, along with the road map, when provided with continuous frame inputs, ensuring coherence between adjacent frames.}
    \label{fig:temporal}
      \vspace{-1em}
\end{figure*}

Additionally, it is worth noting that, based on our experimental results as shown in Tab.~\ref{tab:panacea}, the key for temporal-based detection models lies in accurately positioning and categorizing objects in each frame; detailed information about objects, such as color and brand, is not crucial. As illustrated in Fig.~\ref{fig:temporal}, when provided with continuous frame inputs, the generated images by PerLDiff ensure that the positions and categories of objects, along with the road map, are consistently aligned with the specified conditions between adjacent frames.

\begin{figure*}[t]
    \centering
      \vspace{-1em}
    \includegraphics[width=\textwidth, keepaspectratio]{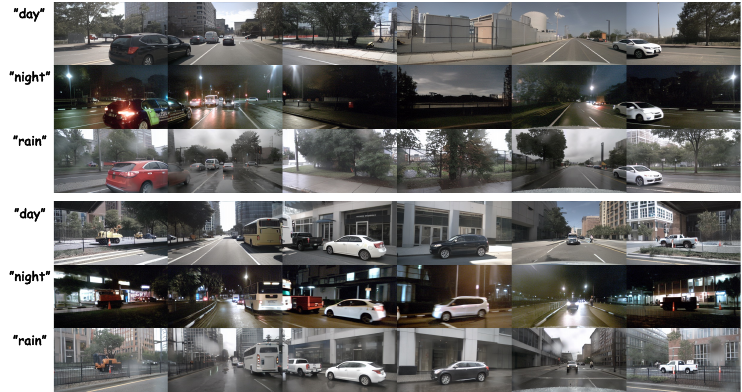}
    \caption{Qualitative visualization on NuScenes demonstrating the effects of Text Cross-attention. From top to bottom: \textit{day}, \textit{night}, and \textit{rain} scenarios synthesized by PerLDiff, highlighting its adaptability to different lighting and weather conditions.}
    \label{fig:sample_wea_1}
      \vspace{-1em}
\end{figure*}

Moreover, Fig.~\ref{fig:sample_wea_1} displays scene alterations by PerLDiff to mimic different weather conditions or times of day, showcasing its versatility in changing scene descriptions. 
Furthermore, as illustrated in Fig.~\ref{fig:fid}, PerLDiff generates background details that do not fully align with those of real images. This discrepancy arises because PerLDiff incorporates prior constraints to ensure accuracy in object detection, which can, in turn, negatively impact the fidelity of the background details.

\begin{figure*}[t]
    \centering
      \vspace{-1em}
    \includegraphics[width=\textwidth, keepaspectratio]{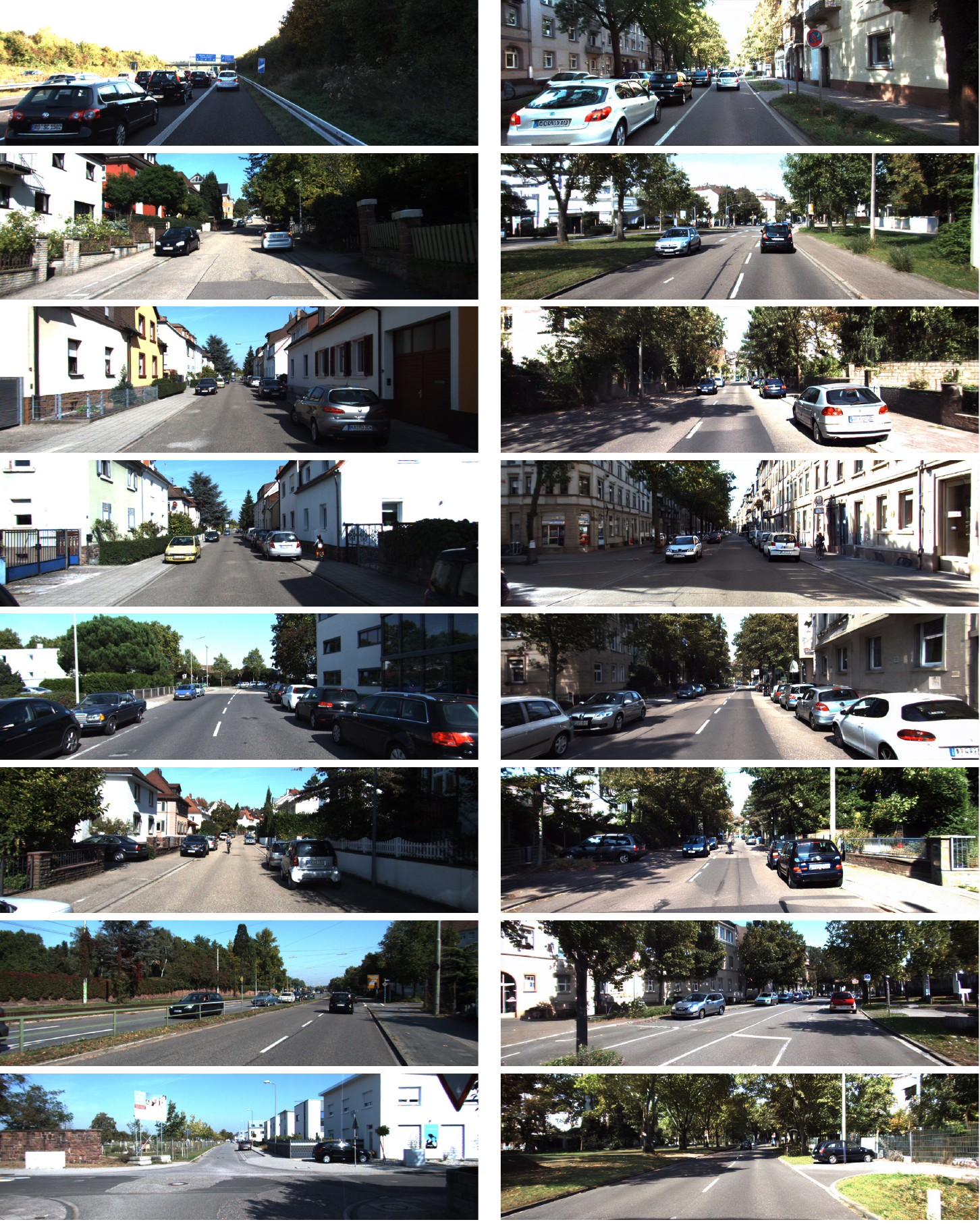}
    \caption{Visualization of street view images generated by our PerLDiff on KITTI \textit{validation} dataset. We show the ground truth (left) and our PerLDiff (right).}
    \label{fig:kitti_img_2}
      \vspace{-1em}
\end{figure*}

Finally, Fig.~\ref{fig:kitti_img_2} presents samples from KITTI \textit{validation} set, illustrating the application's performance in real-world conditions.

\clearpage


\end{document}